\documentclass[default,iicol]{sn-jnl}

\usepackage{algorithm}
\usepackage{algorithmicx}
\usepackage{algpseudocode} 
\usepackage{multicol}

\usepackage{graphicx} 
\usepackage[utf8]{inputenc}
\usepackage{float} 
\usepackage{graphicx} 
\usepackage{array}
\usepackage{lmodern}
\usepackage{anyfontsize}

\usepackage{silence}
\usepackage{longtable}
\usepackage{listings}
\usepackage{subcaption}
\usepackage{cleveref}
\usepackage{cuted}
\usepackage{amsmath}
\usepackage{silence}
\usepackage{mdframed}
\usepackage{multirow}

\jyear{2024}%

\theoremstyle{thmstyleone}%
\theoremstyle{thmstyletwo}%

\theoremstyle{thmstylethree}%

\begin{document}

\title[Article Title]{Experimental Evaluation of Machine Learning Models for Goal-oriented Customer Service Chatbot with Pipeline Architecture}
\markboth{}{} 


\author[1]{\fnm{Nurul Ain Nabilah} \sur{Mohd Isa}}\email{nurulainx96@gmail.com}
\author[2]{\fnm{Siti Nuraishah} \sur{Agos Jawaddi}}\email{aishahagos96@gmail.com}
\author*[1,2]{\fnm{Azlan} \sur{Ismail}}\email{azlanismail@uitm.edu.my}

\affil[1]{\orgdiv{School of Computing Sciences, College of Computing, Informatics and Mathematics}, 
\orgname{Universiti Teknologi MARA (UiTM)}, 
\orgaddress{\city{Shah Alam}, \postcode{40450}, \state{Selangor}, \country{Malaysia}}}

\affil*[2]{
\orgdiv{Institute for Big Data Analytics and Artificial Intelligence (IBDAAI)}, 
\orgname{Kompleks Al-Khawarizmi,  Universiti Teknologi MARA (UiTM)}, 
\orgaddress{\city{Shah Alam}, \postcode{40450}, \state{Selangor}, \country{Malaysia}}}

\abstract{
Integrating machine learning (ML) into customer service chatbots enhances their ability to understand and respond to user queries, ultimately improving service performance. However, they may appear artificial to some users and affecting customer experience. Hence, meticulous evaluation of ML models for each pipeline component is crucial for optimizing performance, though differences in functionalities can lead to unfair comparisons. In this paper, we present a tailored experimental evaluation approach for goal-oriented customer service chatbots with pipeline architecture, focusing on three key components: Natural Language Understanding (NLU), dialogue management (DM), and Natural Language Generation (NLG). Our methodology emphasizes individual assessment to determine optimal ML models. Specifically, we focus on optimizing hyperparameters and evaluating candidate models for NLU (utilizing BERT and LSTM), DM (employing DQN and DDQN), and NLG (leveraging GPT-2 and DialoGPT). The results show that for the NLU component, BERT excelled in intent detection whereas LSTM was superior for slot filling. For the DM component, the DDQN model outperformed DQN by achieving fewer turns, higher rewards, as well as greater success rates. For NLG, the large language model GPT-2 surpassed DialoGPT in BLEU, METEOR, and ROUGE metrics. These findings aim to provide a benchmark for future research in developing and optimizing customer service chatbots, offering valuable insights into model performance and optimal hyperparameters.


\thispagestyle{empty} 
}




\maketitle


\section{Introduction}
\label{sec:introduction}

Concerns have been raised about the quality of customer service in e-commerce, particularly regarding long waiting times and irrelevant responses in live chat interactions~\cite{Adamopoulou2020-fp}. To overcome these issues, chatbots have been used to support customers 24 hours a day. Currently, there are two types of chatbot: rule-based and artificial intelligent (AI)-based~\cite{meshram2021conversational}. The rule-based chatbots use pre-defined sets of rules to answer users' queries meanwhile AI-based chatbots are trained to recognize specific keywords and patterns, which helps in responding to the users' queries. A rule-based chatbot can speed up the process of answering queries, but it is limited to answering frequently asked questions (FAQ). Even though the FAQ chatbot excels in guiding customers through initial problem diagnosis and offering around-the-clock service, its limitations become evident after a few questions, necessitating the involvement of traditional customer service methods. Due to that, an AI-based chatbot that can automate the conversational process is needed. In contrast with a rule-based chatbot, the AI-based chatbot involves the use of Natural Language Processing (NLP) and Machine Learning (ML) algorithms that provide the chatbot with self-learning capabilities. Meanwhile, the ML algorithms that are often considered are deep neural networks (DNN)~\cite{nuruzzaman2018survey} which have been trusted to solve problems involving conversational AI~\cite{simhadri2023effectiveness, verma2023synthesized, chakraborty2023sentiment}. Nevertheless, there is a shortcoming when using AI chatbots, as some users may find the interactions artificial or impersonal, and there is a risk that the chatbots may not fully understand or respond to more complex or nuanced input which eventually affects customer experience and loyalty to the company. Therefore, evaluating the effectiveness of chatbots is crucial to gauge their performance. Experimental evaluation offers a viable approach, allowing researchers to gather empirical evidence across different aspects (e.g. scenarios and metrics). It facilitates identifying strengths and weaknesses through iterative improvements and benchmarking against existing chatbot solutions.



On the other hand, we have identified a group of research endeavors that proposed experimental evaluation for chatbots. They commonly came up with solutions for three reasons. The first reason is to evaluate the performance of the chatbot from a specific aspect, including its architectural component~\cite{keerthana2021evaluating} and the naturalness of the conversation~\cite{atiyah2019evaluation}. The second reason is to evaluate the impact of chatbot approaches toward user experience in which the approaches are conversation types, interaction mechanisms~\cite{haugeland2022understanding}, chatbot suggestions~\cite{gao2021evaluating}, chatbot modalities~\cite{ciechanowski2019shades}, and response conditions~\cite{diederich2019design}. The third reason is to evaluate the suitability of the chatbot in conducting particular tasks such as collaborative working~\cite{ren2020collaborative} and facilitate learning~\cite{akula2024evaluating}. However, among the studies, efforts for evaluating the performance of a customer service chatbot are hardly found, and only one study has considered evaluating the chatbot, from a usability perspective instead of a functional one.  

Meanwhile, to evaluate the functional aspect of a chatbot properly, we first need to consider the type of chatbot and its system architecture. Chatbots can be either task-oriented (also known as goal-oriented), designed for specific goals in short conversations within a limited domain, or non-task-oriented, designed for long, entertaining conversations~\cite{hussain2019survey}. There are also two types of chatbot architectures: end-to-end and pipeline~\cite{chen2017survey}. The end-to-end architecture requires the dialogue to be formulated as a sequence-to-sequence problem, whilst, the pipeline architecture is composed of three components: Natural Language Understanding (NLU), Dialogue Management (DM), and Natural Language Generation (NLG). The goal-oriented chatbot is typically implemented with pipeline architecture~\cite{chen2017survey}. In this study, we propose experimental evaluation for the goal-oriented customer service chatbot with pipeline architecture. Specifically, our aim is to evaluate the chatbot's performance from a functional perspective. Within the evaluation process, we have selected six ML-based models to serve as building blocks of the chatbot components. The goal is to determine the model with the best performance, as well as the key optimal hyperparameters of each model.

In summary, the principal contributions of this study are: 
\begin{itemize}
   \item a methodology to evaluate and optimize candidate models for the key components (Natural Language Understanding, Dialogue Management, and Natural Language Generation) of a customer service chatbot utilizing ML models as presented in Section \ref{sec:experimental-evaluation-design}; 
   \item an analysis to establish the optimal hyperparameter configurations for these models, aiming to enhance chatbot performance, with the results presented in Section \ref{sec:experiments-results};
   \item an evaluation of model performance across chatbot components in simulated customer service scenarios to provide insights into their practical application and effectiveness, discussed in further detail in section \ref{sec:experiments-results}.
\end{itemize}

The remainder of this paper is organized as follows. Section~\ref{sec:background} provides the fundamental background that motivates this research. Section~\ref{sec:relatedwork} compares the existing related works against the proposed evaluation method. Section~\ref{sec:dataset} introduces the dataset used to evaluate the proposed method. Section~\ref{sec:design-development} details the proposed experimental evaluation design and development of the chatbot components. Section~\ref{sec:experiments-results} describes the experimental setup and the analysis results. Section~\ref{sec:discussion} discusses the overall findings of this research. Finally, Section~\ref{sec:conclusion} concludes the paper and recommends future research.

\section{Background}\label{sec:background}
In this section, we explain three fundamental concepts that drive this research: goal-oriented customer service chatbots, chatbots with pipeline architecture, and ML models for pipeline architecture.

\subsection{Goal-oriented Customer Service Chatbots}\label{sec:goal-oriented-cs-chatbots}
Goal-oriented chatbots (also known as task-oriented) are created for specific tasks and designed to engage in brief conversations within a limited scope. In contrast, non-goal-oriented chatbots excel in mimicking human conversation, engaging users in casual chitchat across diverse topics ~\cite{chen2017survey}. The goal-oriented chatbot is usually designed to deal with specific scenarios such as the booking service for hotels or accommodations, product order placement, scheduling events and so on~\cite{hussain2019survey}. The chatbot focuses on helping the user achieve a specific goal and does not have the capability to answer questions related to general knowledge or trivia questions. 

On the other hand, integrating chatbots into e-services has emerged as a promising strategy to enhance customer service nowadays~\cite{chung2020chatbot}. Generally, customer service involves offering information and support to a service provider's clients to enhance users-providers connection and boost company revenue, or to give users the necessary assistance and information they need~\cite{folstad2019chatbots}. Meanwhile, conversational commerce concurrently provides a platform for customers and technology-driven brand representatives to engage in dialogue and transactions, bridging the gap between humans and computers~\cite{Lim2022-ex}. As a result, the presence of customer service chatbots empowers customers to receive personalized recommendations and interact with brands in a humanized environment. A survey identified five customer-related functions employed on a chatbot to achieve two objectives~\cite{misischia2022chatbots}. For the first objective, the functions employed are related to three aspects: interaction, entertainment, and problem-solving. The second objective involves functions related to trendiness and customization aspects. Besides that, there is also a survey that compiled recent studies that have employed the five customer-related functions and identified that customer service chatbots can still build positive customer relationships even having limited communication with the customers~\cite{chung2020chatbot}. 

In line with these findings, our study specifically focuses on developing a customer service chatbot that emphasizes the problem-solving aspect within the restaurant domain. By narrowing the scope in this way, we aim to ensure that the chatbot effectively addresses customer needs while maintaining a personalized and engaging interaction.



\subsection{Chatbot with Pipeline Architecture}\label{sec:chatbot-pipeline-architecture}

\begin{figure}[!ht]
  \centering
  \includegraphics[width=\linewidth]{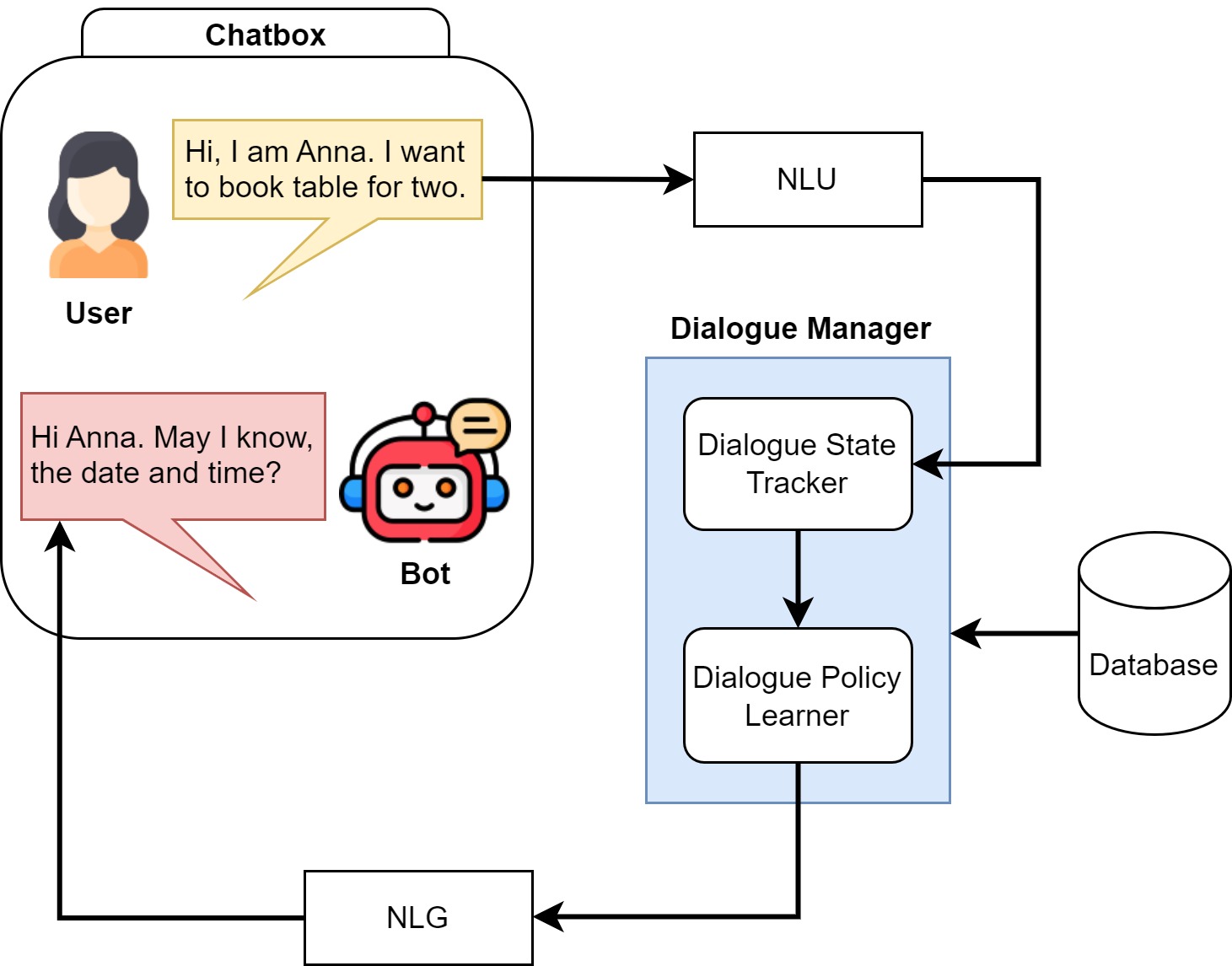}
  \caption{Dialogue flow for a chatbot system with DRL-based dialogue manager}
  \label{fig:illustration}
\end{figure}

Goal-oriented chatbots are commonly built using a pipeline architecture, as outlined in a survey in~\cite{chen2017survey}. This architecture comprises four main components: natural language understanding (NLU), dialogue state tracking (DST), dialogue policy learning (DPL), and natural language generation (NLG). Occasionally, the DST and DPL are combined into a single component known as the dialogue manager (DM). Meanwhile, understanding how each component works is crucial for designing and building effective chatbots that meet their goals such as providing an accurate understanding of user inputs, maintaining coherent conversation states, making informed decisions, as well as, generating natural and relevant responses. Figure~\ref{fig:illustration} illustrates the dialogue flow in the pipeline for a customer service chatbot which was adapted from the traditional pipeline presented in~\cite{chen2017survey}. 

The NLU component is the one that will receive the user utterance and often performs two tasks, namely, intent detection and slot filling to detect the user's intention and assign the words in the sentences with semantic labels, respectively~\cite{weld2022survey}. The semantics will be used for further processes such as information gathering, question answering, dialogue management, and request fulfillment. Meanwhile, the DM component will make use of the chat history which is generally expressed as slot-value pairs to predict the dialogue state and decide the next action that should be taken by the dialogue agent~\cite{dai2020survey}. The first sub-component is the DST which estimate the user's goal at every turn of the dialogue by making use of a classic state structure called semantic frame~\cite{chen2017survey}. A hierarchical semantic frame concept have been introduced to represent the meaning within spoken utterances where the highest level is the domain, then followed by intent, slots, and query~\cite{weld2022survey}. The DST maps the semantic into a suitable state from the state space of the chatbot system which hold the information needed for the chatbot to decide on how to answer the user's questions~\cite{brabra2021dialogue}. On the other hand, the second sub-component which is the DPL utilized the output of DST to determine appropriate dialogue action for the chatbot. Both sub-components traditionally adopt rule-based methods to perform the aforementioned tasks which cause limitations to the chatbots including poor scalability, insufficient tagged data, and low training efficiency~\cite{dai2020survey}. The last component of the chatbot is the NLG component which converts the abstract dialogue action into natural language surface utterance by mapping the semantic symbols input to it to an intermediary form such template structure before converting the intermediary form into the final response through surface realization~\cite{santhanam2019survey}. 

For our experimental evaluation, we are conducting separate assessments of each component's performance. Since each component serves different functions, it would not be equitable to directly compare their performance. Thus, we are evaluating them individually to ensure a more accurate analysis of their effectiveness.

\subsection{ML Models for Pipeline Architecture}\label{sec:MLmodel-pipeline-architecture}

The chatbots have evolved from rule-based to artificial intelligence (AI)-based by considering the NLP and ML methods in the implementation~\cite{meshram2021conversational}. With the implementation of NLP, humans, and bots are able to interact. Meanwhile, with ML methods, the chatbots are now occupied with additional capabilities: self-learning. The self-learning helps the chatbots to train themselves to recognize keywords and patterns to help them provide responses to the users' queries. Additionally, by utilizing the previous interaction with the users, human intervention is less likely required to produce responses for future interaction.  On the other hand, deep learning (DL) is a subset of ML methods that is often considered to solve problems involving conversational AI that which related to extracting the meaning from the input and generating an output as response~\cite{nuruzzaman2018survey}. 

Besides, DL, RL is another subset of ML that has been considered recently to improve the quality of interactions by optimizing dialogue strategies besides enabling the chatbot to handle complex inputs better through continuous learning~\cite{Lin2023-yv}. Regardless of the RL-based methods chosen to develop the DM, it still aims to find the optimal policy that can maximize the expected cumulative reward, however, it struggles to effectively explore and learn optimal policies in an environment with large state and action spaces as it only depends on the lookup table~\cite{fenjiro2018deep}. Therefore, DRL has been considered to overcome this issue while recurrent neural networks (RNNs) and deep Q-Networks (DQNs) are often paired together for natural language processing (NLP)~\cite{Uc-Cetina2023-ce}. RNNs play a key role in processing dialogue history, extracting pertinent details to update the dialogue state in a way that has improved the precision and effectiveness of dialogue systems. Meanwhile, function approximation via neural networks is often used to represent the policy, allowing dialogue systems to handle large state and action spaces. 

Recent studies have commonly employed machine learning models to develop key components of chatbots. For the NLU component, the ML models that have been assessed are BERT~\cite{chen2019bert, natarajan2020semantic, ezen2020comparison, huggins2021practical}, LSTM~\cite{ezen2020comparison}, and Bi-LSTM~\cite{natarajan2020semantic, huggins2021practical}. Findings indicate that BERT can lead to high accuracy in intent classification and slot filling tasks~\cite{chen2019bert}, while LSTM performs better when the chatbot is trained with limited data~\cite{ezen2020comparison}. Methods used in recent studies to build the DRL-based DM component include DQN~\cite{Mukund2019ConvoBotA, nishimoto2022enhancing}, Deep Dyna-Q~\cite{zhang2022dnn}, and Double Actor-Critic (DAC), which is a combination of Double DQN (DDQN) and actor-critic~\cite{saffari2023actor}. Their findings showed that a DQN-based chatbot outperforms a simpler dialogue agent in terms of success rate and reward return, while the combination of DDQN and actor-critic methods achieved higher success rates in task completion. Finally, for the NLG component, existing research has implemented GPT-2~\cite{budzianowski2019hello, andreas2021comparative}, DialoGPT~\cite{nyberg2020response, das2022conversational}, BART, and T5~\cite{andreas2021comparative}. GPT-2 has proven to maintain good performance compared to the original GPT model~\cite{budzianowski2019hello}, while in another finding, DialoGPT performs better than GPT-2~\cite{nyberg2020response, das2022conversational} in terms of generating text that is closer to human-generated dialogue. Thus, we chose to consider LSTM, Bi-LSTM, and BERT for the NLU component, and DQN and DDQN for the DM component. We opted for DDQN over actor-critic to compare against DQN, as both DDQN and DQN share the same fundamental structure of value-based RL. Finally, for the NLG component, we chose to compare DialoGPT and GPT-2 instead of BART and T5, both of which are models derived from BERT.



\section{Related Work} \label{sec:relatedwork}

Current studies that propose experimental evaluation for chatbots commonly exist for three major reasons: to evaluate the performance of the chatbot from a specific aspect~\cite{keerthana2021evaluating, atiyah2019evaluation}, the impact of chatbot approaches toward user experience~\cite{haugeland2022understanding, gao2021evaluating, diederich2019design, ciechanowski2019shades}, and the suitability of the chatbot in conducting particular tasks~\cite{ren2020collaborative, akula2024evaluating}. As for the first reason, a study evaluated the performance of the chatbot from its architectural aspect which involved the author implementing the DM component of the chatbot with three different DRL algorithms and assessing the chatbot's accuracy in answering the user questions~\cite{keerthana2021evaluating}. Meanwhile, another study evaluates the naturalness of the conversation between a personal chatbot and the user which involves assessing the quality of the responses based on the degree of user satisfaction~\cite{atiyah2019evaluation}. For the second reason, studies have evaluated five categories of chatbot approaches: conversation types, interaction mechanisms~\cite{haugeland2022understanding}, chatbot suggestions~\cite{gao2021evaluating}, chatbot modalities~\cite{ciechanowski2019shades}, and response conditions~\cite{diederich2019design}. \citet{haugeland2022understanding} had implemented a chatbot with two types of conversation (task-led or topic-led conversation) and two interaction mechanisms (button or free-text) followed by analyzing the impact from a usability perspective. \citet{gao2021evaluating} have utilized ParlAI~\cite{miller2017parlai} to build four types of chatbots with different capabilities of providing suggestions and evaluating them based on the overall quality, English fluency, and humanlikeness. \citet{ciechanowski2019shades} examined the phenomenon of the uncanny valley in user interactions with various chatbot modalities, such as text-based or avatar, by analyzing psychophysiological factors like heart rate and facial muscle activity. \citet{diederich2019design} implement chatbot with two types of response conditions, namely, control (have present answers) and treatment (have no preset answers) conditions and evaluate the user satisfaction metrics. For the third reason, a study in \cite{ren2020collaborative} has compared the suitability of the SOCIO chatbot with the Creately online tool in terms of facilitating collaborative modeling activity by assessing four usability aspects (efficiency, effectiveness, satisfaction, and quality). Meanwhile, another study evaluated user satisfaction when using the chatbot-based workshop to facilitate remote learning sessions~\cite{akula2024evaluating}.

Existing studies have primarily focused on experimental evaluations of chatbots for general conversation~\cite{diederich2019design, keerthana2021evaluating, ciechanowski2019shades}, knowledge-based interaction~\cite{gao2021evaluating}, personalized tasks~\cite{atiyah2019evaluation}, collaborative work~\cite{ren2020collaborative}, facilitated learning~\cite{akula2024evaluating}, and customer service~\cite{haugeland2022understanding}. However, while these studies have addressed usability aspects, there remains a gap in evaluating customer service chatbots specifically for performance, particularly from an architectural perspective. Our study aims to fill this void by proposing an experimental evaluation approach that assesses the performance of customer service chatbots in terms of intent classification, slot-filling accuracy, the effectiveness of DRL agents in training dialogue management policies, and the capability of large language models to generate diverse responses. Unlike previous work that has focused on usability~\cite{haugeland2022understanding}, our approach aims to delve deeper into the technical aspects of chatbot functionality. Furthermore, we compare and evaluate multiple ML-based models for each chatbot component to identify the most effective models and determine the key hyperparameters influencing their performance.


\section{Dataset} \label{sec:dataset}

The dataset chosen for this study is MultiWOZ 2.2~\cite{budzianowski2018large}, which has been retrieved from the Huggingface website. It is a large-scale, fully annotated corpus of natural human-human conversations that are used for building service agents. It is rich in annotations such as goal, metadata, and dialogue act, along with user and system utterances. These annotations facilitate the use of ML techniques to develop a chatbot. 

The dataset is about the user, as a tourist, converses with the system on service-related topics across multiple domains. Each dialogue in the MultiWOZ dataset consists of a goal, database, and dialogue turns. The goal is defined by the service domain (e.g. hospital, hotel) and the slots. The slots are divided into informable, requestable, and book slots. Informable slots represent user constraints, requestable slots hold additional information that the user wants to obtain, and book slots are used to reserve a place recommended by the system.

In the MultiWOZ dialogues, there are a total of 13 intents including welcome, inform, request, and bye, 24 slots that capture specific information in the dialogues like address, area, price range, and more, as well as a collection of 1482 data points encompassing various types of information such as restaurant names, the day of the week, and other details relevant to the domains in the dataset. Here, each dialogue is a dictionary with keys that also include `turns'. The `turns' key contains a list of turns, where each turn is a dictionary with keys such as `speaker' and `utterance', which represent the speaker of that turn (either `0', user, or `1', system) and the text of the utterance, respectively. The uttered exchange between the two speakers can be captured by extracting the `speaker' and `utterance' values for each turn.

The MultiWOZ dataset has been used in various studies to fine-tune models for task-oriented dialogue systems. This dataset has also been used in benchmark tasks such as Dialogue State Tracking and Dialogue-Context-to-Text Generation~\cite{zang2020multiwoz}. It is available in the .json format.

\begin{figure}[htbp]
  \centering
  \includegraphics[width=\linewidth]{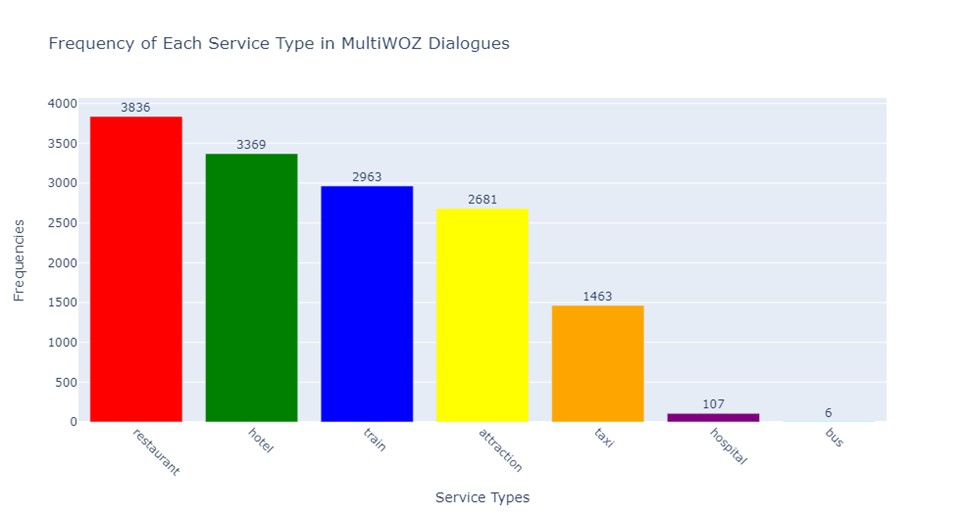}
  \caption{The frequency of dialogue for each customer service domain in MultiWOZ dialogue data.}
  \label{fig:Fig1} 
\end{figure}

As seen in Figure~\ref{fig:Fig1}, the restaurant service type makes up a bulk of the MultiWOZ dialogue data, containing a variety of restaurant-related exchanges that encompass reservations, cuisine preferences, location inquiries, and customer service interactions. Given this substantial representation, the development of this project’s chatbot and its components will be centred on the restaurant domain.

Thus, for all model training purposes, the MultiWOZ dataset has been filtered to include only dialogues about the restaurant domain. The names of user intents and chatbot action types within the dataset have also been consolidated into broader categories after removing domain-specific prefixes via the process described in Algorithm~\ref{Algo:mapping}:

\begin{algorithm}
\caption{Mapping Dialogue Act Types}
\label{Algo:mapping}
\begin{algorithmic}[1]

\State \textbf{Input:} MultiWOZ dialogue with turns containing domain-specific act types
\State \textbf{Output:} MultiWOZ dialogue with simplified act types

\State Define a mapping from original, domain-specific act types to more simplified act types (e.g. Restaurant-Request to Request)

\For{each turn in the dialogue}
    \State Retrieve original act types from the turn
    \State Map original act types to simplified act types
    \State Update the turn with simplified act types

    \If{the turn contains span information with act types}
        \State Retrieve original act types from span information
        \State Map these original act types to simplified act types
        \State Update the span information with simplified act types
    \EndIf
\EndFor

\end{algorithmic}
\end{algorithm}


\section{Design and Development}\label{sec:design-development}

In this section, we present the proposed experimental evaluation design and describe the development of the chatbot components.

\subsection{Experimental Evaluation Design}\label{sec:experimental-evaluation-design}
In order to comprehensively assess the entire chatbot system in terms of its functionality, we have designed a two-phase experiment. The first phase focuses on identifying the optimal hyperparameters, while the second phase focuses on identifying the ML that contributes to the best performance for the chatbot.

For the first phase, we conducted hyperparameter optimization (HPO), a process aimed at finding the right combination of hyperparameters to enhance the performance of ML models and yield better results~\cite{Shekhar2021-yo}. This phase comprises three key steps: identifying the tool for HPO, defining a feasible range of configurations for the hyperparameters, and determining the optimal hyperparameters based on the provided configuration. Conventional methods for conducting HPO include grid search~\cite{hsu2003practical} and random search~\cite{bergstra2012random}, known for their ability to navigate high-dimensional spaces effectively and converge quickly, respectively. However, these methods frequently face runtime challenges in production environments and may lack robust performance, issues that Bayesian optimization algorithms effectively tackle. Several Python libraries now support automated HPO which is based on the Bayesian algorithms, that includes Sequential Model-based Algorithm Configuration (SMAC), HyperOpt, Optunity, and Optuna. SMAC excels in selecting optimal parameters across large configuration spaces and reduces optimization time by leveraging predictive models of runtime~\cite{hutter2011sequential}. However, it may struggle with extremely high-dimensional search spaces due to internal model complexity. HyperOpt, on the other hand, offers versatility in optimizing various types of variables and supports parallel implementation through platforms like Apache Spark and MongoDB, making it advantageous for large-scale problems~\cite{bergstra2013hyperopt}. Nevertheless, it only provides limited search algorithms compared to other options (i.e. random search, Tree-structured Parzen Estimator (TPE), and adaptive TPE). In contrast, Optunity's main advantage is its user-friendly design that simplifies the hyperparameter optimization process while ensuring compatibility and ease of integration across different environments~\cite{claesen2014easy}. As Optunity avoids dependencies on large packages to facilitate use in non-Python environments, this may lead to a slower optimization process. Nevertheless, the emergence of Optuna~\cite{akiba2019optuna} addresses the drawbacks effectively. Optuna not only delivers a user-friendly interface for analyzing optimization outcomes but also boasts scalability, accommodating anything from individual trial optimizations to extensive experimentation. It utilized efficient sampling methods like TPE and Covariance Matrix Adaptation Evolution Strategy (CMA-ES), which can handle larger and more complex search spaces more effectively. Additionally, it can handle various independent sampling and relational sampling methods, as well as, depending on big Python packages such as NumPy for better optimization performance and provide more features. Furthermore, Optuna provides advanced features like pruning to stop unpromising trials early, which can significantly speed up the optimization process. Due to the advantages of Optuna, we decided to utilize this tool in this study, to perform HPO during the evaluation process. On the other hand, the details for the second and third steps have been described in Section~\ref{sec:experiments-results} and Section~\ref{sec:analysis-results}, respectively.

The second phase involves three steps: selecting ML models for each chatbot component along with suitable performance metrics, implementing the selected ML methods for each component development, and analyzing their performance based on the chosen metrics. The ML models selected for each component are based on the findings discussed in Section~\ref{sec:MLmodel-pipeline-architecture} and the detail justification has been discussed in Section~\ref{sec:development-of-component}. Specifically, LSTM and BERT models are chosen for the NLU component while DQN and DDQN are selected for the DM component, and GPT-2 and DialoGPT are chosen for NLG. Performance evaluation metrics vary for each component. For the NLU component, accuracy is used for intent detection evaluation, while recall, precision, and F1 score assess slot and slot value detection. In the DM component, performance evaluation relies on three metrics: average reward per episode, average number of dialogue turns per conversation, and task success rate. The task success rate measures the percentage of dialogues where the dialogue manager agent achieves the simulated user's goal by matching its final state, including gathered slots and values, with the user's goal state. As for the NLG component, evaluating large language models (LLMs) necessitates automated metrics, hence we considered BLEU, METEOR, and ROUGE in our experiment. Details for the subsequent steps are elaborated in Section~\ref{sec:analysis-results}.

\subsection{Development of Chatbot Components}\label{sec:development-of-component}
\begin{figure}[htbp]
  \centering
  \includegraphics[width=\linewidth]{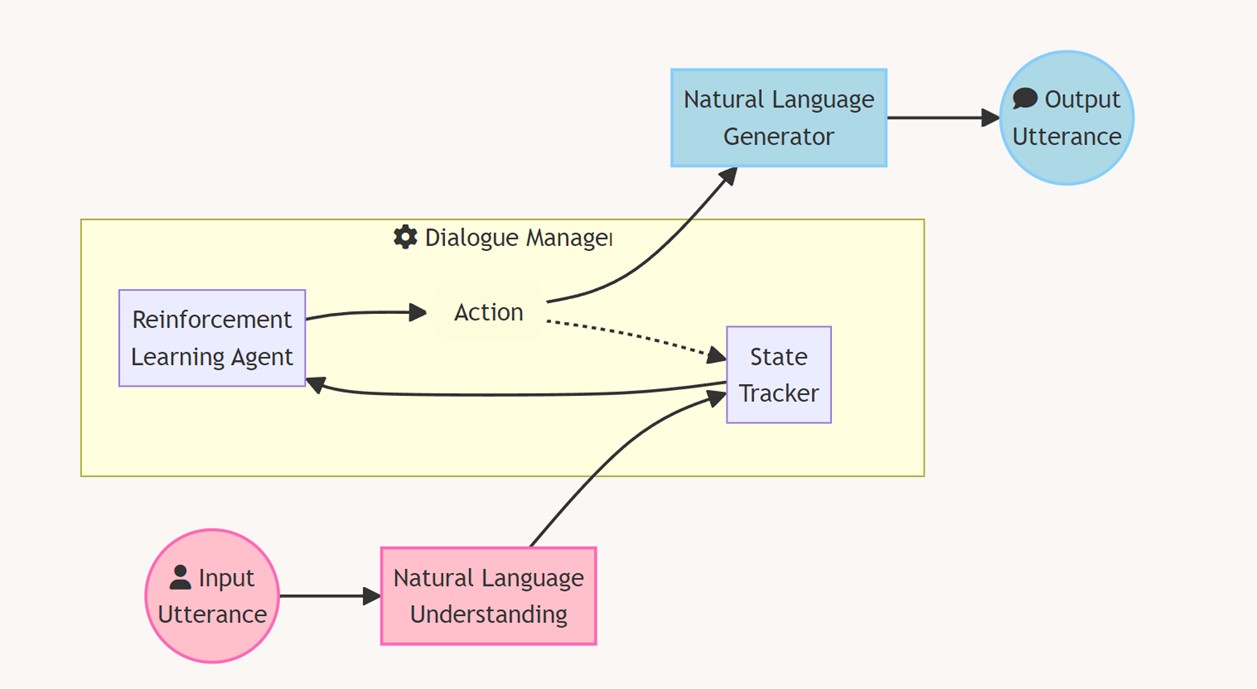}
  \caption{Interaction between the key chatbot components}
  \label{fig:Fig2} 
\end{figure}

Three key chatbot components need to be developed to complement the steps taken for the second phase of our experimental evaluation design. The components are NLU, DM, and NLG while Figure~\ref{fig:Fig2} illustrates the interaction between the components.

\subsubsection{NLU Component}
The first component is the NLU, involved in comprehending user intents by extracting the semantics information from user utterances as slots to be processed further for information gathering, question answering, dialogue management, request fulfillment, and others~\cite{weld2022survey}. This component is a fusion of an ML-based model and a variety of natural language processing methods, including the tokenization of user speech into words and sub-words, (beginning-inside-outside) BIO-tagging to identify types and boundaries of entities in a speech, and label encoding to convert intents and slot tags into a numerical format. These techniques are all utilized to interpret user inputs accurately. In this component, our attention is primarily on two models, specifically, Long Short-term Memory (LSTM)~\cite{hochreiter1997long} and Bidirectional Encoder Representations from Transformers (BERT)~\cite{devlin2018bert}.

LSTM is a variant of RNN that has been instrumental in progressing complex language processing tasks such as language translation~\cite{torfi2020natural}. The ability of LSTMs to grasp long-term dependencies in data, bypassing the optimization hurdles typically seen in simpler RNNs, is particularly noteworthy. A notable enhancement in LSTM technology is the advent of Bi-directional LSTMs~\cite{graves2005framewise}, which process data in both forward and backward directions, offering a more holistic comprehension of context. The NLU is suitable to be implemented with the Bi-LSTM because of its proficiency in managing long-term dependencies and its bidirectional methodology, enabling it to process text bidirectionally and thereby grasp the context and subtleties of user statements. On the other hand, BERT is a potential model for DL composed of a series of encoder layers that employ self-attention mechanisms to dissect and scrutinize sequences of input tokens~\cite{Rogers2020-br}. For each token within the input sequence, BERT calculates the corresponding key, query, and value vectors that are utilized to construct a weighted representation for each token. These representations undergo processing in several ``heads," enhanced with self-attention, which allows BERT to model dependencies between tokens over long distances. By initially training the model on vast quantities of text data through two semisupervised tasks, namely masked language modeling and next sentence prediction, BERT can be further refined to execute specific tasks. We chose BERT due to its capacity to model long-range dependencies between tokens and its pre-existing training which equips it with the advanced language comprehension necessary to discern the rationale and specifics of a statement.


Before they can be deployed, these two models require training using the MultiWOZ dataset. The training process begins with an initial preprocessing step to identify and classify utterances, intents, and entity slots within user dialogues. A crucial part of this preprocessing involves the application of the BIO tagging scheme, which allows the model to accurately identify and categorize the input data. In this scheme, words in user utterances are labeled as 'B' if they initiate a data segment, 'I' if they extend a segment, or 'O' if they are not part of any segment. To streamline this tagging process, a Python script is created. The script starts by breaking down customer dialogue into its fundamental elements—intent and slot values—and then assigns the relevant BIO tags. It also ensures that any special situations, such as ambiguous slot values, are treated consistently by the models during their training. The tagging process is illustrated by Algorithm~
\ref{Algo:annotating}:

\begin{algorithm}
\renewcommand{\thealgorithm}{2}
\caption{Annotating Slots in User Utterances for NLU Model}
\label{Algo:annotating}
\begin{algorithmic}[2]

\State \textbf{Input:} Utterances, intents, and slots from the MultiWOZ dataset
\State \textbf{Output:} Tokenized utterances with BIO-tagged slot annotations
\State Initialize empty lists for tokenized utterances, intents, and slot annotations

\For{each entry in data}
    \State Extract `utterance`, `intent`, and `slots` from the entry
    \State Tokenize the `utterance`
    \State Initialize slot annotation lists with 'O' tags for each token

    \For{each slot and corresponding value in slots}
        \If{value indicates a request slot}
            \State Tokenize the slot name
            \State Annotate the corresponding utterance tokens with 'B-' and 'I-' tags for request slots
        \Else
            \State Tokenize the slot value
            \State Annotate the corresponding utterance tokens with 'B-' and 'I-' tags for inform slots

            \If{slot requires special handling (e.g., 'moderately priced' as a synonym for 'moderate' in pricerange)}
                \State Handle special cases by identifying and annotating synonyms
            \EndIf
        \EndIf
    \EndFor

    \State Append the processed utterance, intent, and annotations to respective lists
\EndFor

\end{algorithmic}
\end{algorithm}

After tagging, the utterances would then be tokenized to prepare them for ingestion by the models. While LSTM employs standard tokenization, BERT uses a more granular subword tokenization technique~\cite{Rogers2020-br}. Following tokenization, the tokens and slot tags are converted into numerical indices, and sequences are padded to a fixed length for uniform input size to address the challenge of variable-length input sequences. 

The architecture of the LSTM model begins with an embedding layer that converts input text into dense vector representations~\cite{yao2018improved}, followed by the namesake LSTM layers: for intent classification, a single LSTM layer is used while for slot filling tasks, Bi-directional LSTM layers are used to better match the task’s increased complexity. These two layers represent separate branches for handling different types of slots, 'inform' and 'request.' The output of these LSTM layers is then fed into dense layers with a softmax activation function that categorizes each token into a specific slot type and classifies the overall intent of the utterance. On the other hand, the BERT model utilizes the small-size pre-trained BERT model, `bert-base-uncased’~\cite{devlin2018bert}. The input data is first processed by the BERT tokenizer and then fed into the BERT layers. The BERT model output is then used to create different branches for intent detection and request slot-filling tasks. For intent classification, the output corresponding to the [CLS] or classify token (which represents the entire sentence) is passed through a dense layer with softmax activation to classify the user's intent. For slot filling, the output embeddings for all tokens are processed through a time-distributed dense layer, again with softmax activation. Like its LSTM counterpart, the model also has separate branches for 'inform' and 'request' type slots. 

\subsubsection{DM Component}
The second component, which is also the chatbot's core function relies on the DM component that helms a decision-making system similar to Markov Decision Processes (MDPs). In this setup, a DRL agent acts as the decision-maker, choosing the next conversation step based on the current dialogue context. The goal is to maximize the expected total reward, aligning with the chatbot's aim to fulfill the user's intent effectively, just like how optimal policies work in MDPs. Meanwhile, the state tracker or DST maintains an evolving representation of the conversation's context in a way similar to state transitions in MDPs. It keeps track of past interactions and updates the perceived context after each user input and chatbot response, mirroring how states change in an MDP after each action. DQN and DDQN are chosen as candidates for the chatbot's dialogue manager for distinct reasons. DQN offers a compelling choice due to its famous utilisation of a deep convolutional neural network to represent the Q function so it can learn directly from high-dimensional sensory inputs. This can be advantageous in processing complex conversational contexts. On the other hand, DDQN may mitigate potential overestimations of action values~\cite{Agostinelli2018-mg}. This overestimation concern can affect performance in dialogue management tasks, so DDQN’s usage of two separate Q-network copies may provide a more accurate and stable decision-making approach during user-chatbot interactions.

For this study, each DRL agent employs a sequential model with two fully connected layers, utilising ReLU activation functions. An adjustable hidden size parameter sets the first layer's size while the second is half of this first layer. The final output layer uses a linear activation function. The model is compiled with Mean Squared Error (MSE) loss and uses the Adam optimiser with a predefined learning rate.

These agents, adaptations of Max Brenner’s GO-Bot-DRL and a simplified rendition of MiuLab’s TC-Bot, distinguish themselves primarily through their network configurations: DQN employs a single network for both action selection and Q-value estimation while DDQN utilises a dual-network approach, separating action selection and Q-value estimation processes.

\subsubsection{NLG Component}
The third component, the NLG is responsible for bringing the chatbot's interactions to life. Once the DM component decides on the next action, it will be this component's role to transform this decision into a human-like reply. This transformation is done by generating text word by word where each word is selected based on the probabilities conditioned not only on the preceding words but also on the input from the DM component. This generative process, guided by the weights within its large language model that has been fine-tuned to the MultiWOZ dataset, will produce coherent responses to customer queries.

The first candidate for the NLG component is GPT-2, an autoregressive language model introduced by OpenAI in 2019~\cite{solaiman2019release}. It uses a transformer-based neural network architecture with 12-48 layers, trained on 40 GB of internet text data. GPT-2 is able to generate fluent, coherent text that closely matches human writing by predicting the next word in a sequence based on the previous words—so fluent, it has even raised questions about potential for malicious use of large, unchecked language models once released. Meanwhile, the second candidate is DialoGPT, an open-domain pre-trained conversational response generation model released by Microsoft in 2019. It extends the GPT-2 architecture to dialogue modelling by training it on 147 million conversation-like exchanges from Reddit spanning 2005-2017~\cite{zhang2019dialogpt}. GPT-2 is chosen as a candidate for its known ability for a wide range of natural language generation tasks like summarisation, translation, question answering, and conversation while DialoGPT is chosen for its ability to produce diverse, relevant, and contentful responses to conversational input.

For this study, the selected candidate models will be used in their most compact form, approximately 500MB in size and encompassing 117 million parameters. Still, before their deployment, GPT-2-small and DialoGPT-small require fine-tuning with the MultiWOZ dataset. For the fine-tuning to happen, the dataset has to undergo pre-processing. Since the NLG part of the chatbot is meant to convert an action semantic frame into a sentence in natural language, only system dialogues—examples of possible actions taken by a chatbot and their corresponding utterances—would be extracted from MultiWOZ.

To facilitate this transformation, the data comprising extracted dialogues and their action semantic frames must be reformatted into a serialized form. A serialization process within the MultiWOZ context will transform an example system utterance with its corresponding action frame into a linear, easily parsed string. The incorporation of a unique delimiter token ``\textless\textbar\textbar\textgreater'' serves to segment different components within the serialised action-utterance pair for the model to comprehend and process. This format is important when it comes to presenting the semantic frame-to-utterance conversion in a manner that is sequentially coherent and conducive to learning large language models.

This string of action-utterance pairs will then be used to fine-tune both models. After pre-processing, the dataset is partitioned into a training and testing split, with a 90:10 ratio, respectively. By feeding the model dialogue that has been tokenized by GPT2Tokenizer, the model’s parameters will be optimized based on the dialogue to minimize the loss function, ensuring that it learns to generate responses similar to those in the dataset.

Altogether, whenever a user interacts with the chatbot, the NLU will process the input and extract necessary information like intents and entities. This information is then passed to the DM component, which uses its learned policy to decide the best action to take in the current state. The decided action is then passed to the NLG component to generate a natural language response that will soon be presented to the user.



\section{Experiments and Results} \label{sec:experiments-results}

To evaluate the effectiveness of targeted ML-based models in a goal-oriented customer service chatbot, we have designed the experiments to first focus on finding optimal hyperparameters for each candidate model using the Optuna tool~\cite{akiba2019optuna} before comparing the performance of each optimized model in pairs across the system's three main components: NLU, dialogue management, and NLG. Two following research questions (RQs) were proposed to drive the evaluation process which the detailed answer discussed in Section~\ref{sec:discussion}:
\begin{itemize}
    \item RQ1: What are the hyperparameters that significantly influence the models selected for each goal-oriented customer service chatbot's component (NLU, DM, NLG) and their optimal values?
    \item RQ2: Which selected models can give out the best performance for each component of the goal-oriented customer service chatbot?
\end{itemize}

\subsection{Experiment Setup} \label{sec:experimental-setup}
In this section, we describe the configuration settings for each experiment which have been renamed as Experiment 1, Experiment 2, and Experiment 3. Each of the experiments is associated with the evaluation of the components for NLU, dialogue management, and NLG, respectively.

For the NLU component, we aimed to identify the best-performing model for intent detection and slot filling. We evaluated two candidates: a joint LSTM model with an LSTM branch for intent detection and dual bi-LSTM branches for \textit{inform} and \textit{request} slot detection, and a \textit{bert-base-uncased} BERT model. Training and testing for both models were conducted on user utterances sourced from the MultiWOZ restaurant domain dataset, employing a 70:30 split ratio for training and testing purposes. Additionally, the models are compiled using Adam optimizer with specified learning rates and \textit{clipnorm} = 1.0. In the hyperparameter optimization setup for the NLU candidate models, the parameters optimized for the BERT model are \textit{learning rate} and \textit{batch size}, while for the LSTM model, the parameters are \textit{learning size}, \textit{batch size}, \textit{embedding dimensions}, and \textit{numbers of LSTM units}. Table~\ref{table:nlu-hyperparameters} shows the configuration set for the hyperparameters. The number of epochs in the BERT model is fixed to 4 to balance between allowing enough iterations for the model to learn from the data and constraining the training time to a reasonable limit, considering the complexity and large size of the BERT. In contrast, the LSTM model underwent a more prolonged learning period with 16 epochs to capitalize on its sequential data processing strengths. On the other hand, metrics used to evaluate the performance of the models include accuracy for intent detection, and recall, precision, and F1 score. 

\begin{table}[ht]
\centering 
\begin{tabular}{|p{0.8cm}|p{1.5cm}|p{1cm}|p{1.6cm}|p{1cm}|}
\hline
\textbf{Model} & \textbf{Learning Rate} & \textbf{Batch Size} & \textbf{Embedding Dimensions} & \textbf{LSTM Units} \\
\hline
BERT & 1e-5 to 1e-3 (log scale) & 16, 32, 64 & N/A & N/A \\
\hline
LSTM & 1e-4 to 1e-2 (log scale) & 16, 32, 64 & 64, 128, \newline 256 & 64, 128, \newline 256 \\
\hline
\end{tabular}
\caption{Hyperparameters for NLU models BERT and LSTM.}
\label{table:nlu-hyperparameters}
\end{table}

Meanwhile, for the DM component, we aimed to evaluate and determine the best performer between DQN and DDQN models in terms of managing dialogues to achieve user goals. The hyperparameter optimization setup for these models was designed to ensure a fair comparison between the two models, with both models sharing the same hyperparameters-\textit{learning rate}, \textit{batch size}, \textit{hidden layer size}, and \textit{initial epsilon value}—as well as search space. For each of the trials held, both models would be trained with 10,000 episodes of conversations with the user simulator. Additionally, the DM component is evaluated based on its average reward per episode, the average number of dialogue turns per conversation, and the task success rate. The task success rate is defined as the percentage of dialogues in which the DM's agent successfully completes the simulated user's goal. To determine if the agent has successfully completed the user's goal, the agent's final state (i.e., the gathered slots and values) is compared to the user's goal state. If the agent's final state matches the user's goal state, then the chatbot has successfully completed the task.

\begin{table}[ht]
\centering
\begin{tabular}{|p{1cm}|p{1.2cm}|p{1cm}|p{1cm}|p{1cm}|}
\hline
\textbf{Model} & \textbf{Learning Rate} & \textbf{Batch Size} & \textbf{Hidden Layer Size} & \textbf{Initial Epsilon} \\
\hline
DQN \& DDQN & 1e-4 to 1e-2 (log scale) & 64, 128, 256 & 60, 80, 100, 120, 140 & 0.1 to 0.5 \\
\hline
\end{tabular}
\caption{Hyperparameters for DM models DQN and DDQN.}
\label{table:dialogue-manager-hyperparameters}
\end{table}

Lastly, for the NLG component, we aimed to assess which model between GPT-2 and DialoGPT could transform semantic action frames into natural, conversational language more effectively. These models were subjected to a consistent hyperparameter space focusing on the learning rate and batch size to guarantee comparability as two variants of the same large language model architecture. A comparatively low learning rate search space (between 1e-5 to 1e-4) is used based on the recommendations by~\citet{ameri2021cybert}. Each model was trained across the fixed amount of 10 epochs on the semantic frame to chatbot utterance translation dataset in each of its respective trials, balancing between an ample duration of data exposure and the constraints of time as well as computation typical of large language models. Meanwhile, evaluating large language models (LLMs) for NLG would require automated metrics such as BLEU, METEOR, and ROUGE~\cite{sai2022survey}. Traditional human evaluations, while accurate, are expensive and time-consuming. These three metrics are designed to automate the evaluation process by comparing the generated text against a set of reference texts.

\begin{table}[ht]
\centering
\begin{tabular}{|c|c|c|}
\hline
\textbf{Model} & \textbf{Learning Rate} & \textbf{Batch Size} \\
\hline
GPT-2 & 1e-5 to 1e-4 (log scale) & 4, 8, 16 \\
\hline
DialoGPT & 1e-5 to 1e-4 (log scale) & 4, 8, 16 \\
\hline
\end{tabular}
\caption{Hyperparameters for NLG models GPT-2 and DialoGPT.}
\label{table:nlg-hyperparameters}
\end{table}

\subsection{Result Analysis} \label{sec:analysis-results}

In this section, we discuss the result analysis according to Experiments 1, 2, and 3 from two perspectives: the optimized hyperparameter and the model performance.

\subsubsection{Experiment 1: Analysis on NLU component}

\begin{figure*}[!htb]
\centering
  \begin{subfigure}{0.7\textwidth}
    \begin{mdframed}
      \centering
      \includegraphics[width=\linewidth]{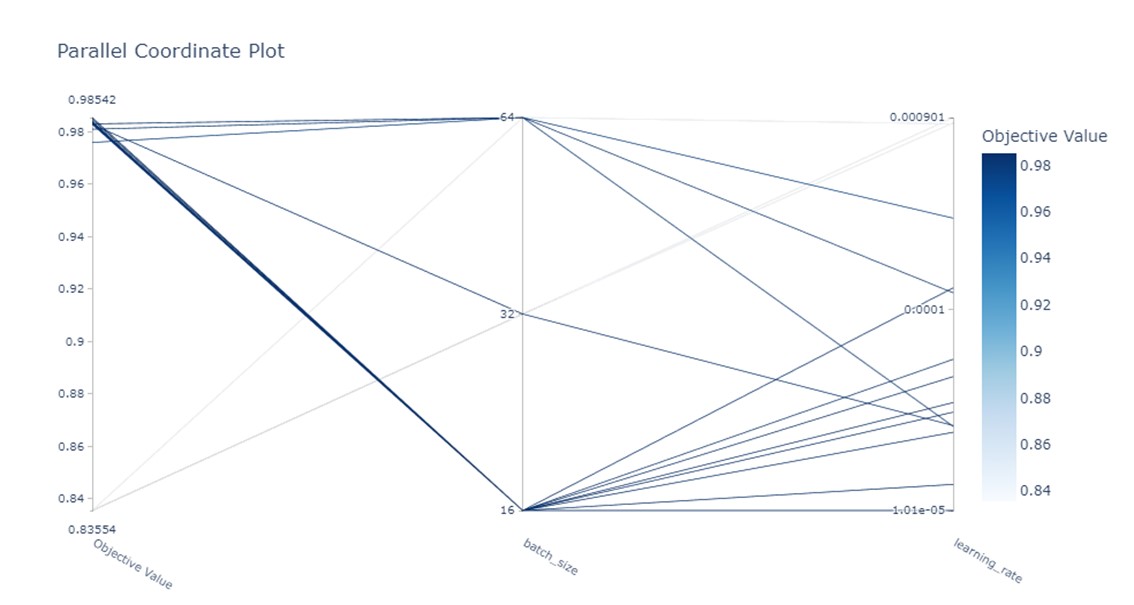}  
    \end{mdframed}
    \caption{Optimized hyperparameter values for BERT}
    \label{fig:r1}
  \end{subfigure}
  \hfill
  \begin{subfigure}{0.7\textwidth}
    \begin{mdframed}
      \centering
      \includegraphics[width=\linewidth]{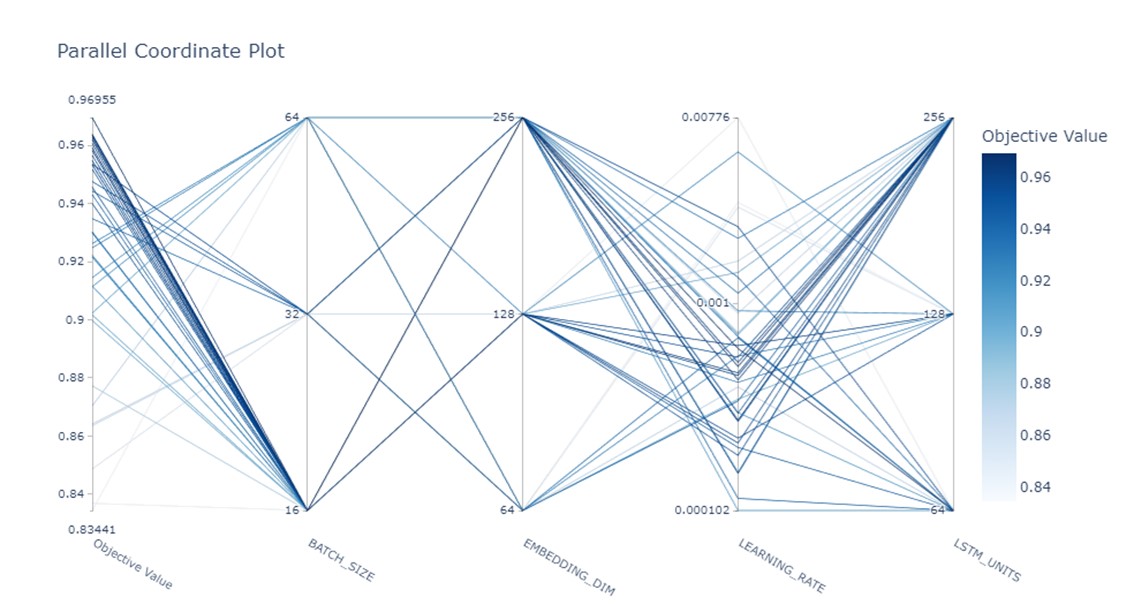}  
    \end{mdframed}
    \caption{Optimized hyperparameter values for LSTM}
    \label{fig:r2}
  \end{subfigure}
  \caption{Results for the experiment on natural language understanding's component}
  \label{fig:experiment-result-1-a}
\end{figure*}

Figures~\ref{fig:r1} and~\ref{fig:r2} illustrate the parallel coordinate plot to visualize the correlation between multiple hyperparameters within BERT and LSTM models, respectively.

\begin{figure}[!htb]
\begin{mdframed}
  \centering
  \includegraphics[width=\linewidth]{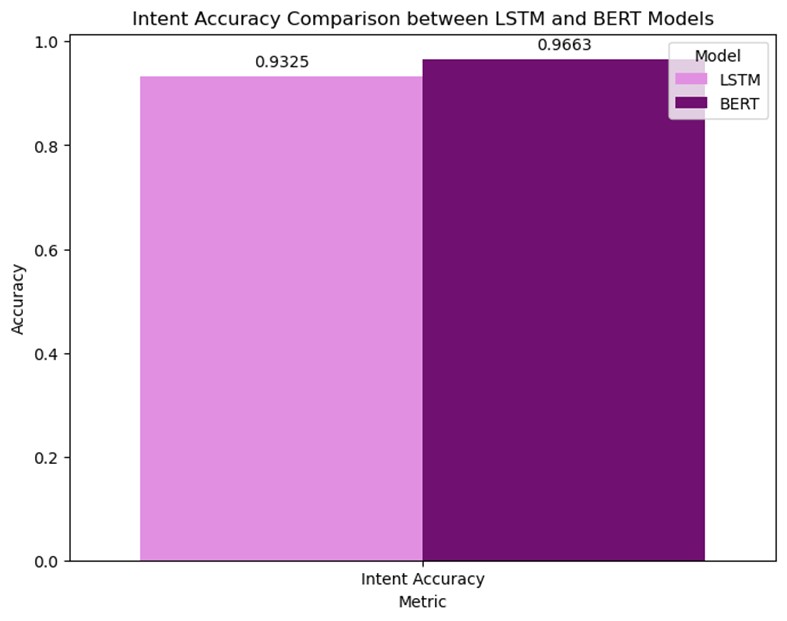} 
  \end{mdframed}
  \caption{Comparison of intent detection accuracy between the LSTM and BERT models}
  \label{fig:experiment-result-1-b}
\end{figure}

Based on our observation, the BERT model reached its best performance when the batch size was small and the learning rate closer to the 1e-5 end of the log scale. These choices fit the small-sized single-domain dataset where high learning rates lead to untimely suboptimal convergence and small batch sizes will reduce overfitting. Meanwhile, based on the importance score, \textit{learning rate}(0.93) has a more significant impact on the model performance compared to \textit{batch size}(0.07). On the other hand, the LSTM model can perform at its best with the smallest batch size (i.e. 16) and the learning rate is below 0.001. However, embedding dimensions and the number of LSTM units seem to have no single dominating value. In terms of importance scores, the learning rate (0.47) is the most influential hyperparameter, followed by batch size (0.34), embedding dimension (0.14), and number of LSTM units (0.05). This underscores the critical role of the learning rate in the LSTM's learning process, significantly impacting how quickly and effectively the model converges to an optimal solution. Batch size also plays a crucial role by affecting the model's training stability and speed. Although the embedding dimension and the number of LSTM units do influence performance, their impact is less pronounced compared to the learning rate and batch size in this optimization scenario.

\begin{table*}[!htb]
\centering
\caption{Comparison of precision, recall and F1-score for conversation-relevant slot and slot value detection between LSTM and BERT}
\label{tab:comparison-nlu-model} 
\begin{tabular}{p{0.18\textwidth} p{0.08\textwidth} p{0.11\textwidth} p{0.08\textwidth} p{0.08\textwidth} p{0.11\textwidth} p{0.08\textwidth} p{0.08\textwidth}}
 \hline 
\textbf{Slot Type} & \textbf{Model Type} & \textbf{LSTM Precision} & \textbf{LSTM Recall} & \textbf{LSTM F1-Score} & \textbf{BERT Precision} & \textbf{BERT Recall} & \textbf{BERT F1-Score}\\
  \hline
{B-AREA} & {Inform} & {0.96} & {0.99} & {0.97} & {0.81} & {0.8} & {0.81}\\
{B-PRICERANGE} & {Inform} & {0.94} & {0.97} & {0.96} & {0.81} & {0.85} & {0.83}\\
{B-NAME} & {Inform} & {0.59} & {0.36} & {0.44} & {1} & {0.04} & {0.07}\\
{I-PRICERANGE} & {Inform} & {0.94} & {0.96} & {0.95} & {0.76} & {0.77} & {0.77}\\
{I-NAME} & {Inform} & {0.53} & {0.33} & {0.41} & {0.71} & {0.11} & {0.19}\\
{I-FOOD} & {Inform} & {1} & {0.41} & {0.58} & {0.82} & {0.24} & {0.38}\\
{B-FOOD} & {Inform} & {0.93} & {0.94} & {0.94} & {0.72} & {0.8} & {0.76}\\
{B-POSTCODE} & {Request} & {1} & {0.84} & {0.91} & {0.87} & {0.91} & {0.89}\\
{B-FOOD} & {Request} & {1} & {0.33} & {0.5} & {0} & {0} & {0}\\
{B-PHONE} & {Request} & {0.99} & {1} & {0.99} & {0.9} & {0.92} & {0.91}\\
{B-ADDRESS} & {Request} & {0.99} & {1} & {1} & {0.9} & {0.92} & {0.91}\\\hline

\end{tabular}
\end{table*}

Further evaluation of the performance can be referred to in Figure~\ref{fig:experiment-result-1-b} and Table~\ref{tab:comparison-nlu-model}. Figure~\ref{fig:experiment-result-1-b} compares the intent detection accuracy of the two models. The result shows that BERT outperforms the LSTM with a difference of approximately 0.0338. The average accuracy of the models is calculated as shown in Equation~\ref{eq:avg-accuracy}, where $\alpha$ is the accuracy of intent prediction, $\beta$ is the \textit{inform} slot detection and $\gamma$ is the \textit{request} slot detection. The reason for BERT's superior performance is its transformer-based architecture~\cite{devlin2018bert} that processes words in context more effectively than LSTM's sequential approach~\cite{hochreiter1997long} as its pre-training on vast text data gives it a deep understanding of language nuances. However, BERT's higher accuracy comes with greater computational demands, something that should be considered for resource-constrained chatbot deployments. Interestingly, in terms of slot filling, LSTM outperforms BERT. Table~\ref{tab:comparison-nlu-model} presents a comparison of LSTM and BERT models for slot and slot value detection on a small dataset of 2963 utterances. Overall, LSTM performs better than BERT in filling the `Inform' and `Request' slots, with values for precision, recall, and F1-score ranging from 0.9 to 1.0. Higher precision values indicate high specificity, while higher recall indicates high sensitivity. Greater F1-scores reflect robustness, showing that the model can effectively detect these slots despite the limited size of the dataset.

\begin{equation}
\text{Average Accuracy} = \frac{\alpha + \beta + \gamma}{3}
\label{eq:avg-accuracy}
\end{equation}



\begin{figure*}[!htb]
\centering
  \begin{subfigure}{0.7\textwidth}
    \begin{mdframed}
      \centering
      \includegraphics[width=\linewidth]{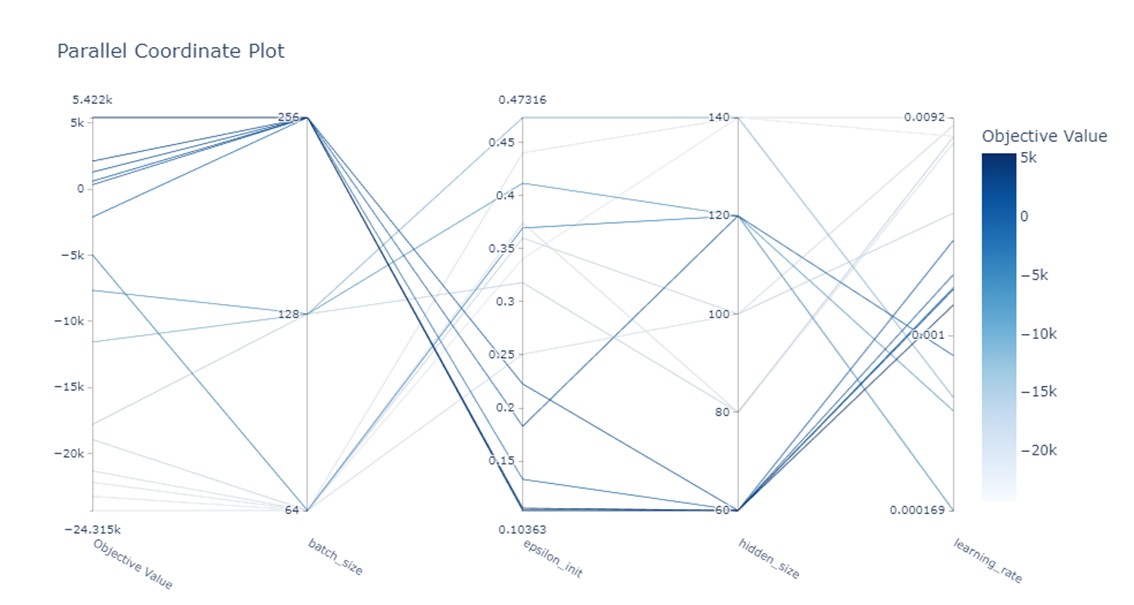}  
    \end{mdframed}
    \caption{DQN}
    \label{fig:r4}
  \end{subfigure}
  \hfill
  \begin{subfigure}{0.7\textwidth}
    \begin{mdframed}
      \centering
      \includegraphics[width=\linewidth]{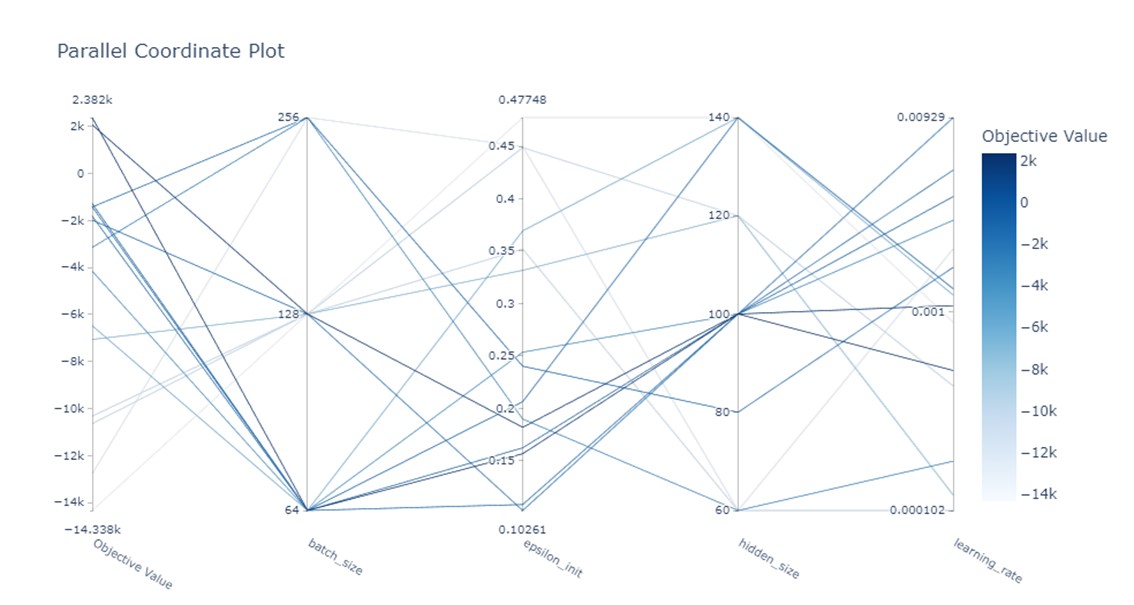}  
    \end{mdframed}
    \caption{DDQN}
    \label{fig:r5}
  \end{subfigure}
  \caption{Results of hyperparameter optimization for DM's component}
  \label{fig:output-bert-lstm}
\end{figure*}

\begin{figure*}[!htb]
\centering
  \begin{subfigure}{0.49\textwidth}
    \begin{mdframed}
      \centering
      \includegraphics[width=\linewidth]{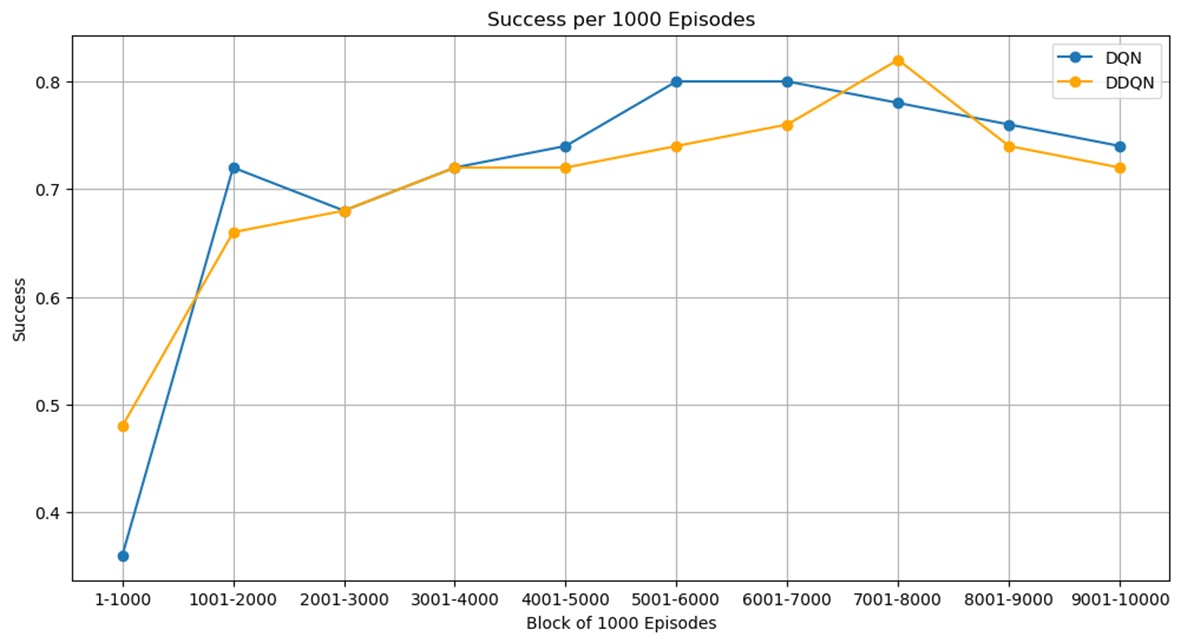}  
    \end{mdframed}
    \caption{Average success rate}
    \label{fig:r6}
  \end{subfigure}
  \hfill
  \begin{subfigure}{0.49\textwidth}
    \begin{mdframed}
      \centering
      \includegraphics[width=\linewidth]{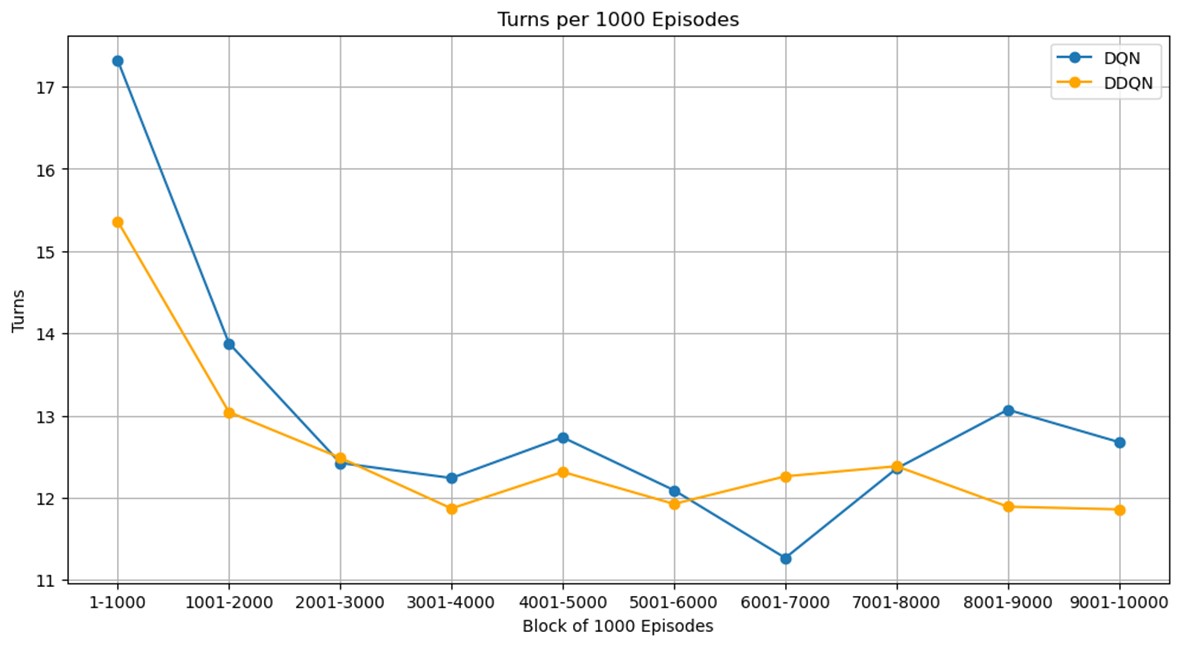}  
    \end{mdframed}
    \caption{Average dialogue turns per conversation }
    \label{fig:r7}
  \end{subfigure}
  \hfill
  \begin{subfigure}{0.49\textwidth}
    \begin{mdframed}
      \centering
      \includegraphics[width=\linewidth]{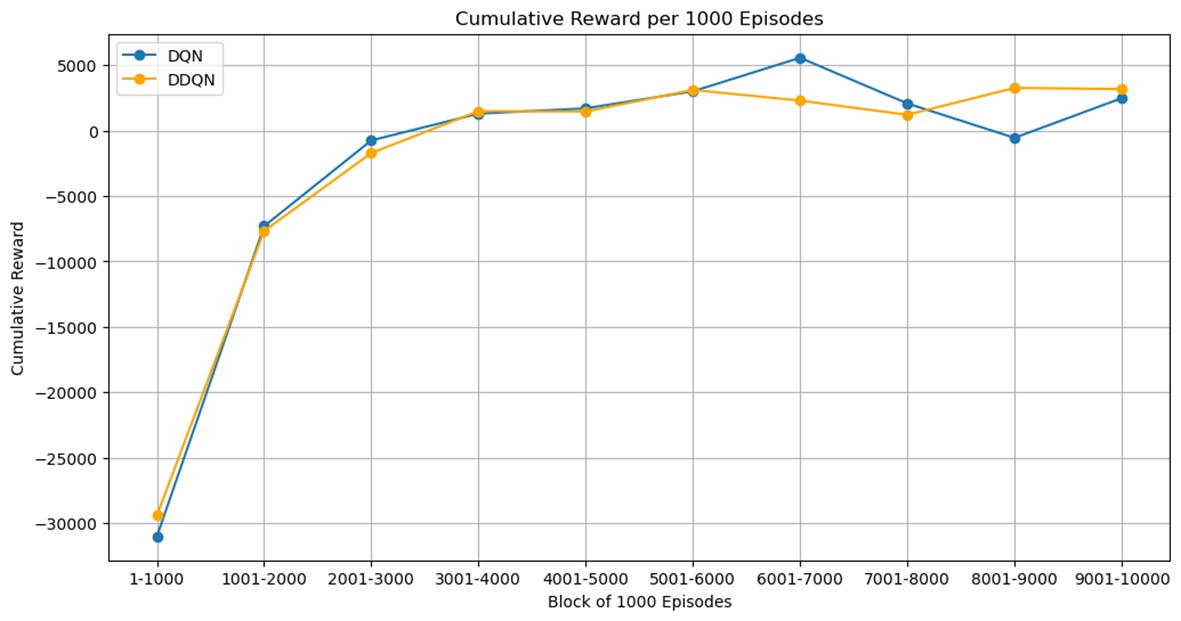}  
    \end{mdframed}
    \caption{Cumulative rewards}
    \label{fig:r8}
  \end{subfigure}
  \caption{Results for the experiment on dialogue manager component involving DQN and DDQN (for every 1000 episodes after 10,000 training episodes)}
  \label{fig:experiment-result-2-a}
\end{figure*}

\begin{figure*}[!htb]
  \begin{mdframed}
  \centering
  \includegraphics[width=0.8\linewidth]{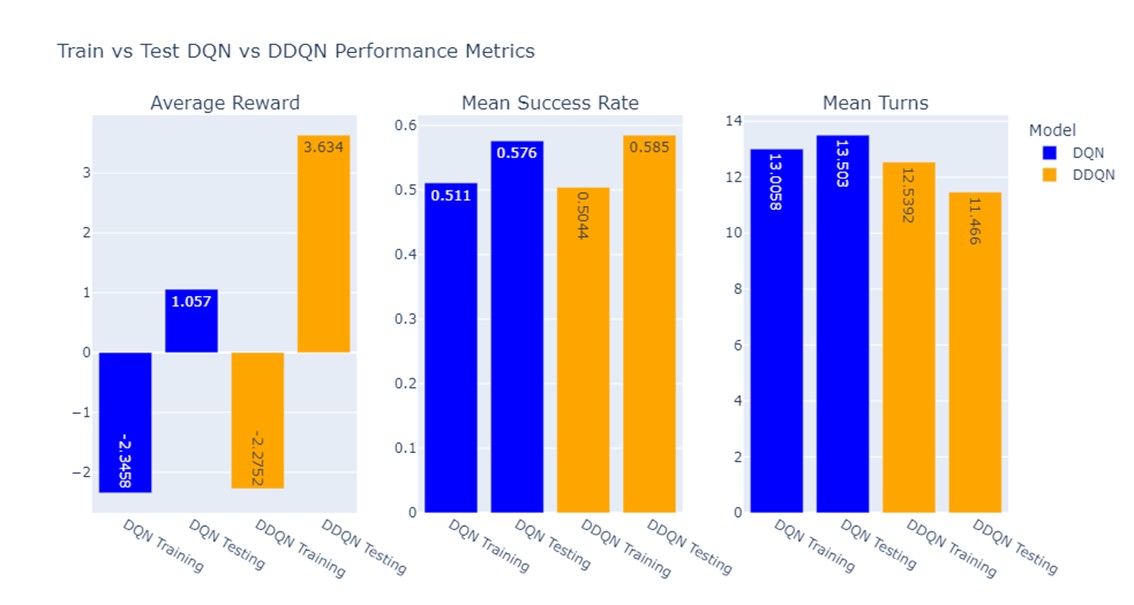}
  \end{mdframed}
  \caption{Comparison of DQN and DDQN performance during training and testing by average rewards, mean success rate and mean number of dialogue turns}
  \label{fig:experiment-result-2-b}
\end{figure*}

\subsubsection{Experiment 2: Analysis on DM component}

Figures~\ref{fig:r4} and~\ref{fig:r5} display the hyperparameters' correlation of DQN and DDQN, which both were investigated using the same hyperparameters. The DQN model is able to reach its best performance when trained with a larger \textit{batch size} and \textit{learning rate} closer to the upper limit of 1e-2. Meanwhile, there are no noticeable trends for the effects of \textit{hidden layer size} and \textit{initial epsilon value} on the model. On the other hand, the DDQN model is able to perform best when the trials with the highest objective values are characterized by lower \textit{initial epsilon values}, especially around the range of 0.1 to 0.2. The remaining parameters (\textit{learning rate}, \textit{batch size}, and \textit{hidden node size}) do not show a clear trend or consistent pattern when it comes to performance.

Additionally, in terms of importance score, the hyperparameter that has the highest significant impact on the performance of the DQN model is the \textit{learning rate}(0.71) followed by \textit{batch size}(0.18) while the \textit{initial epsilon value}(0.06) and \textit{hidden node size}(0.05) are less likely to affect the model's performance. However, for the DDQN model, \textit{initial epsilon value}(0.95) influenced the model performance better than the other hyperparameters: \textit{learning rate}(0.02), \textit{hidden node size}(0.02), \textit{batch size}(0.01).

Figure~\ref{fig:experiment-result-2-a} shows the results collected every 1000 episodes during the training of two models over 10,000 episodes. We analyzed the DM component from three perspectives: average success rate, average dialogue turns per conversation, and cumulative rewards. Several key trends emerge from these training-related charts. Both models exhibit an increasing success rate during training, demonstrating their ability to learn and manage dialogues in a customer service setting. As shown in Figure~\ref{fig:r6}, the DQN model generally surpasses the DDQN model, except between the 6001$^{st}$ and 9000$^{th}$ episodes. However, the DDQN model consistently shows a slightly higher success rate, particularly in the initial training phases, reaching a peak success rate of 82\%. Additionally, both models display a decreasing trend in the number of dialogue turns as training progresses, indicating improved efficiency in resolving issues with fewer interactions as shown in Figure~\ref{fig:r7}. The DDQN model, in particular, demonstrates a steeper decline and stabilizes at a lower number of turns compared to the DQN model. Furthermore, the cumulative rewards for both models increase over time as depicted in Figure~\ref{fig:r8}, showing their ability to maximize rewards and improve dialogue management. The DDQN model exhibits a quicker ascent and achieves a higher cumulative reward by the end of the training.


\begin{figure*}[!htb]
\centering
  \begin{subfigure}{0.7\textwidth}
    \begin{mdframed}
      \centering
      \includegraphics[width=\linewidth]{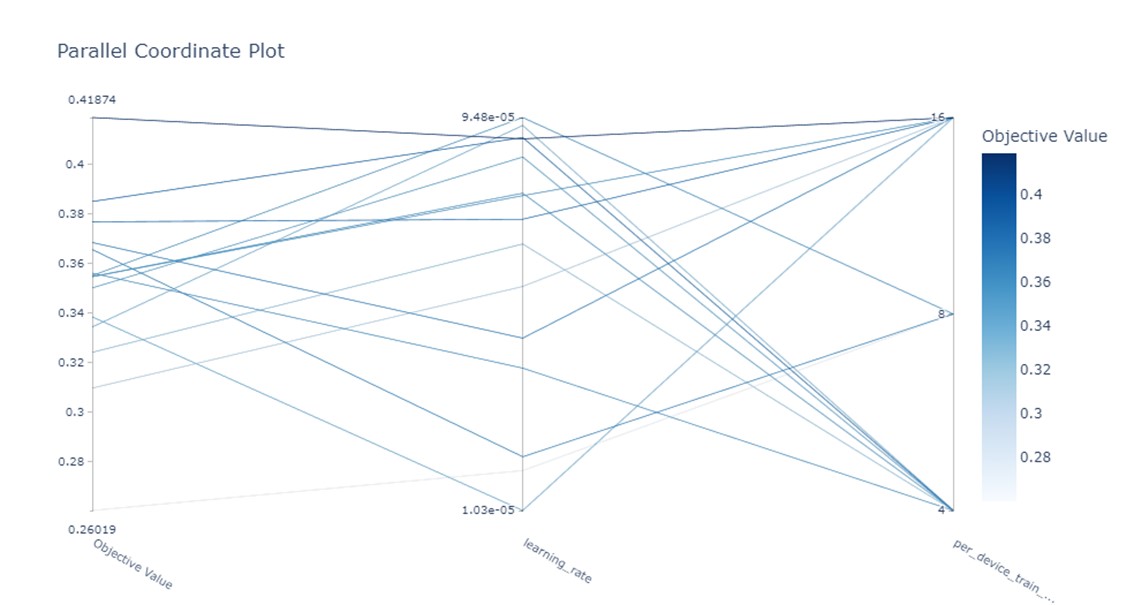}  
    \end{mdframed}
    \caption{Optimized hyperparameter values for GPT-2}
    \label{fig:r10}
  \end{subfigure}
  \hfill
  \begin{subfigure}{0.7\textwidth}
    \begin{mdframed}
      \centering
      \includegraphics[width=\linewidth]{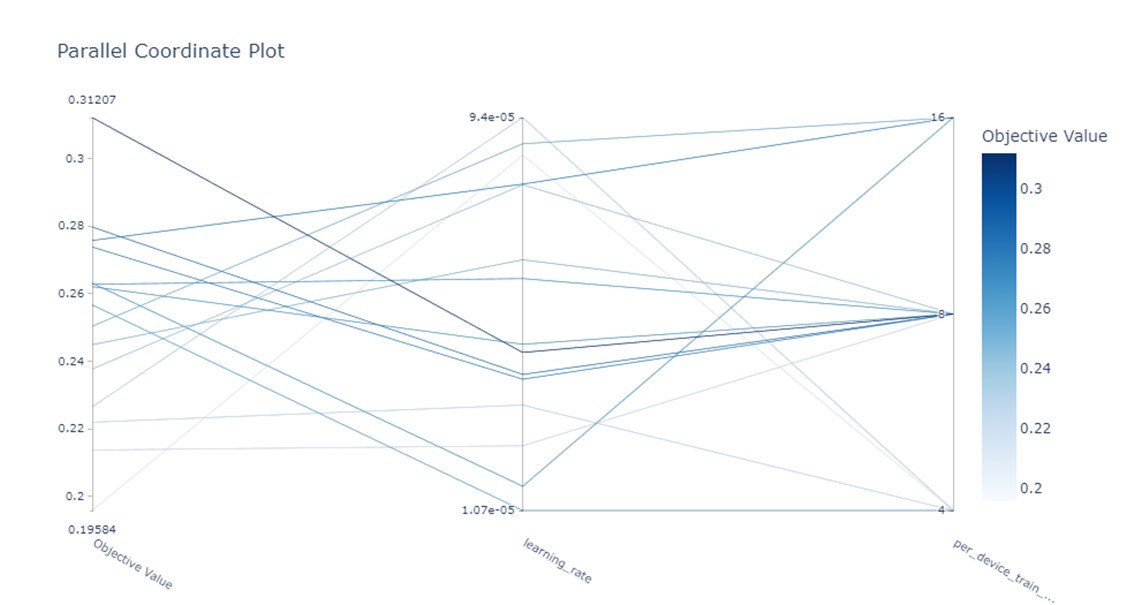}  
    \end{mdframed}
    \caption{Optimized hyperparameter values for DialoGPT}
    \label{fig:r11}
  \end{subfigure}
  \caption{Results for the experiment on natural language generation's component}
  \label{fig:experiment-result-3}
\end{figure*}

To further distinguish which model performs better, we have conducted additional evaluations that assess the performance of both models during the training and testing processes. The results have been illustrated in Figure~\ref{fig:experiment-result-2-b}. We could see significant improvement in these DRL models as the \textit{average reward} experienced a positive increment during the testing process with DDQN (-2.2752 to 3.634) able to get better rewards than DQN (-2.3458 to 1.057). Regarding the \textit{mean success rate}, the DQN model experienced an increase from 0.511 to 0.576 while the DDQN model showed a slight increase from 0.5044 to 0.585. As for \textit{mean turns per conversation}, the DQN model increased from 13.0058 to 13.503 while the DDQN model’s mean turns have reduced significantly from 12.5392 to 11.466. Comparatively, the DDQN model consistently outperforms the DQN model, maintaining stable performance from training to testing. Increasingly higher average rewards, a slight mean success rate improvement, and reduced mean turns signify DDQN's superior adaptability and efficiency in using dialogue strategies for successful outcomes, while DQN lags behind in testing scenarios. Overall, the DDQN model, with its higher rewards, greater success rates and shorter conversations in both train and test stages, emerges as a more suitable choice for a dialogue manager in a customer service chatbot environment, adept at handling varied and new dialogue scenarios.

\subsubsection{Experiment 3: Analysis on NLG component}

Figures~\ref{fig:r10} and~\ref{fig:r11} present the optimized hyperparameters for the models selected for the NLG component, namely, GPT-2 and DialoGPT. Through the observation, GPT-2 can reach its best performance with the largest batch size of 16 and learning rates closer to the higher end of the tested range, around 1e-4. On the other hand, for DialoGPT, a variety of configurations lead to higher objective values, with no single clear trend dominating the results. This means that the DialoGPT model is somewhat flexible in its hyperparameter settings, allowing for a range of values to yield satisfactory results when it comes to text generation. Additionally, the \textit{learning rate} significantly affects the performance of both models with importance scores of 0.72(GPT-2) and 0.52(DialoGPT). Meanwhile, the importance score of the \textit{per-device train batch size} for DialoGPT(0.48) is higher than the one for GPT-2(0.28) and closer to the score of its \textit{learning rate}. It means that the DialoGPT is sensitive to these two hyperparameters. This sensitivity underscores the model's adaptability across diverse conversational contexts. By meticulously adjusting these hyperparameters, the system can better accommodate various linguistic patterns and user interactions, leading to improved performance in novel dialogue scenarios.

\begin{figure}[!ht]
  \begin{mdframed}
  \centering
  \includegraphics[width=\linewidth]{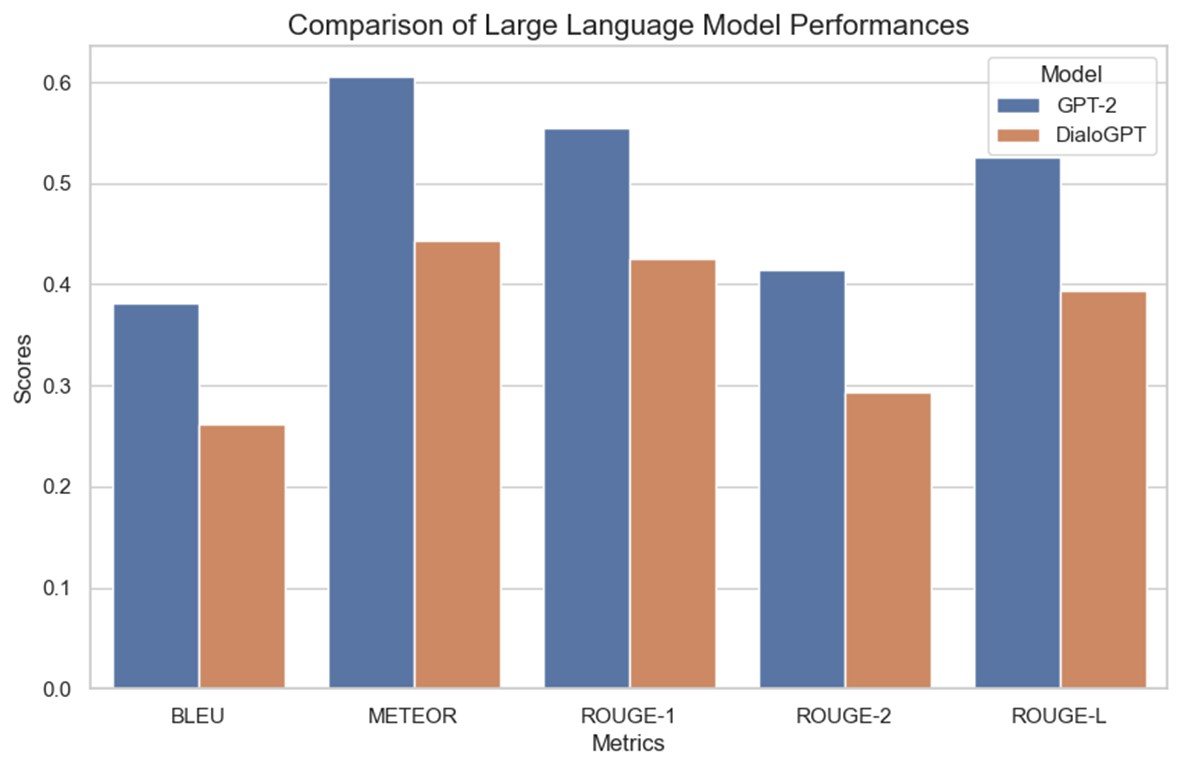}
  \end{mdframed}
  \caption{Comparison of large language model performances}
  \label{fig:experiment-result-3-b}
\end{figure}

Nevertheless, even though the DialoGPT is more flexible than GPT-2, but when it comes to performance, GPT-2 is able to surpass the DialoGPT. The result can be observed in Figure~\ref{fig:experiment-result-3-b}. The GPT-2-small model demonstrated superior performance across all metrics, registering scores for BLEU(0.3814), METEOR(0.6051), ROUGE-1(0.5552), ROUGE-2(0.4141), and ROUGE-L(0.5258). Conversely, the DialoGPT-small model yielded lower scores, with BLEU at 0.2616, METEOR at 0.4428, ROUGE-1 at 0.4249, ROUGE-2 at 0.2937, and ROUGE-L at 0.3937.
These results indicate that GPT-2, with its configured hyperparameters, is more adept at translating semantic action frames into sentences that are accurate enough to be similar to the target references. Specifically, GPT-2’s greater BLUE score indicates better accuracy in mimicking human language while the superior METEOR score means it produces more contextually relevant responses. Additionally, GPT-2's strong ROUGE scores indicate its ability to maintain longer text sequences that match reference texts. This suggests that it can convey information more closely resembling an ideal response.
It may be that GPT-2's training on a broad array of web text has made it a more versatile general language model thus giving it an edge over DialoGPT which, despite specializing in dialogue pairs from Reddit, might not align as closely with varied training data. As the best NLG component, GPT-2's design for wide-ranging text generation tasks makes it more well-suited for crafting translated sentences based on semantic frames, unlike DialoGPT which may be better at conversational responses.


\section{Discussion} \label{sec:discussion}

In this section, we discuss the findings of the evaluation process based on the proposed RQs. We aim to provide insights that could help researchers select potential models to be implemented in their goal-oriented customer service chatbots and select the values of hyperparameters for benchmarking purposes.

Recalling the models we selected in Section~\ref{sec:experimental-evaluation-design}, RQ1 focuses on identifying the hyperparameters that most influenced the performance of the models for each chatbot component (NLU, DM, and NLG). For the NLU component, we chose LSTM and BERT models, with the \textit{learning rate} being the most influential hyperparameter for both. Additionally, LSTM is sensitive to changes in batch size. The optimal \textit{learning rate} values for LSTM and BERT are $6.1 \times 10^{-4}$ and $3.5 \times 10^{-5}$, respectively, while the optimal \textit{batch size} for both models is 16. Other optimal values for LSTM include 256 for the embedding dimension and 64 for the LSTM units. For the DM component, we selected DQN and DDQN to control the conversation flow. DQN's performance is significantly affected by changes in the \textit{learning rate}, whereas DDQN is influenced by the \textit{initial epsilon value}. The optimal hyperparameters for DQN are a learning rate of $1.365 \times 10^{-3}$, a batch size of 256, a hidden size of 60, and an initial epsilon of $1.0577 \times 10^{-1}$. For DDQN, the optimal values are a learning rate of $5.1 \times 10^{-4}$, a batch size of 64, a hidden size of 100, and an initial epsilon of $1.5678 \times 10^{-1}$. For the NLG component, we evaluated GPT-2 and DialoGPT models. The performance of both models is sensitive to changes in the \textit{learning rate}, with optimal values of $8.4027 \times 10^{-5}$ for GPT-2 and $2.561 \times 10^{-5}$ for DialoGPT. Additionally, DialoGPT's performance is significantly influenced by the \textit{per-device train batch size}, with an optimal value of 8, while the optimal batch size for GPT-2 is 16.

Subsequently, after experimenting with the chatbot with specific configurations as explained in Section~\ref{sec:experimental-setup}, we addressed RQ2 by identifying which model provided the best performance for each component. For the NLU component, BERT outperformed LSTM in intent detection accuracy, achieving 96.63\% compared to 93.25\%. This is attributed to BERT's superior contextual processing capabilities, despite its higher computational resource requirements. However, in slot detection, LSTM demonstrated better performance than BERT, particularly in the `Inform' slots. Each model exhibits unique strengths: BERT is ideal for applications prioritizing accuracy in customer service chatbots where high computational resources are acceptable, whereas LSTM is more suitable for deployments with limited computational resources and acceptable accuracy levels in intent detection. For the DM component, DDQN outperformed DQN, achieving higher success rates, fewer dialogue turns, and greater accumulated rewards. This indicates DDQN's enhanced learning efficiency and adaptability in managing dialogues, due to its mechanism of reducing Q-value overestimation. Lastly, GPT-2 excelled in the NLG component, surpassing DialoGPT in all evaluation metrics (BLEU, METEOR, ROUGE). GPT-2's general-purpose nature and training on diverse web text enable it to generate more accurate, contextually relevant responses, making it the preferable model for translating action frames into natural language utterances.

\section{Conclusion} \label{sec:conclusion}

This paper presents an experimental evaluation of a pipeline-driven, goal-oriented customer service chatbot from a functional perspective. The proposed method involves separately evaluating three chatbot components: Natural Language Understanding (NLU), Dialogue Management (DM), and Natural Language Generation (NLG). BERT and LSTM models were chosen for the NLU component, DQN and DDQN for the DM component, and GPT-2 and DialoGPT for the NLG component. Additionally, automated hyperparameter optimization was performed using the Optuna tool, which identified the optimal hyperparameters and their values. Through this evaluation, we determined which machine learning models deliver high performance for each chatbot component and identified the optimal hyperparameter values for each model. The results indicate that for the NLU component, BERT (learning rate: $3.5 \times 10^{-5}$, batch size: 16) excelled in intent detection, while LSTM (learning rate: $6.1 \times 10^{-4}$, batch size: 16) was superior for slot filling. For dialogue management, the DDQN model (learning rate: $5.1 \times 10^{-4}$, batch size: 64) outperformed DQN by achieving fewer turns, higher rewards, and greater success rates. For NLG, the large language model GPT-2 (learning rate: $8.4027 \times 10^{-5}$, batch size: 16) surpassed DialoGPT in BLEU, METEOR, and ROUGE metrics when trained with a small dataset.

Future research can investigate additional important factors that may affect the optimal deep learning model configurations, such as the choice of optimizer, the utilization of dropout and its rates, as well as the exploration of various test-train split and sampling methods. Model performance can also be further enhanced via the incorporation of more sophisticated techniques like transfer learning or ensemble methods. Furthermore, with the rapid progression in conversational AI, the opportunity to incorporate cutting-edge models such as GPT-4 or Mistral 8x7B into customer service is both timely and promising. Another compelling research direction is the real-world deployment of a deep reinforcement learning-based chatbot to actively engage with customers in the real world and to evaluate its performance by using well-established customer service metrics such as Net Promoter Score (NPS), Customer Satisfaction Score (CSAT), and Customer Effort Score (CES). Demonstrating the performance of the chatbot empirically will provide definitive information on the practical effectiveness of the chatbot and user satisfaction.


\backmatter

\section*{Ethics approval and consent to participate}
Not applicable. 

\section*{Consent for publication}
Not applicable.

\section*{Availability of data and materials}
Not applicable.

\section*{Competing interests}
The authors declare that they have no competing interests.

\section*{Funding}
Not applicable.

\section*{Authors' contributions}
Nurul Ain Nabilah Mohd Isa conducted the experiments, analysed the data and wrote the paper; Siti Nuraishah Agos Jawaddi wrote and revised the paper; Azlan Ismail reviewed the experiments, wrote and revised the paper. All authors have read and agreed to the published version of the manuscript.

\section*{Acknowledgement}
We would like to acknowledge the School of Computing Sciences, College of Computing, Informatics and Mathematics, Universiti Teknologi MARA (UiTM) for providing the necessary resources, facilities, and environment that supported our research endeavors.

\bigskip

\bibliographystyle{plainnat}
\bibliography{sn-bibliography}

\clearpage

\end{document}